\documentclass{llncs}
\usepackage[utf8]{inputenc}
\usepackage{amsmath}
\usepackage{amssymb}
\usepackage{gensymb}
\usepackage{graphicx}
\usepackage{booktabs}
\usepackage{tabularx}
\usepackage{textcomp}
\usepackage{hyperref}
\usepackage{nicefrac}
\usepackage{listings}
\usepackage[caption=false]{subfig}
\usepackage[version=0.96]{pgf}
\usepackage[
  separate-uncertainty = true,
  multi-part-units = repeat
]{siunitx}
\usepackage{tikz}
\usepgfmodule{oo}
\usetikzlibrary{arrows,shapes,snakes,automata,backgrounds,petri,calc,arrows,shapes.arrows,arrows.meta}
\usepackage[caption=false]{subfig}
\newcommand{\overbar}[1]{\mkern 1.5mu\overline{\mkern-1.5mu#1\mkern-1.5mu}\mkern 1.5mu}

\usepackage{courier}
\usepackage[caption=false]{subfig}
\usepackage[version=0.96]{pgf}
\usepackage{tikz}

\renewcommand{\eqref}[1]{Eq.~(\ref{#1})}

\DeclareMathOperator{\Sigmoid}{S}
\DeclareMathOperator*{\argmin}{arg\,min}

\newcommand{\minipagetwolayout}[1]{
\captionsetup[subfigure]{labelformat=empty}
\def\imgsize{0.172}
\def\betweenimgsize{0.04}
\begin{minipage}{\betweenimgsize\textwidth}\subfloat[$x_\mathcal{B}$]{~}\end{minipage}
\begin{minipage}{\imgsize\textwidth}\includegraphics[width=\textwidth]{images/#1/flair.png}\end{minipage}
\begin{minipage}{\imgsize\textwidth}\includegraphics[width=\textwidth]{images/#1/t1ce.png}\end{minipage}
\begin{minipage}{\imgsize\textwidth}\includegraphics[width=\textwidth]{images/#1/t1.png}\end{minipage}
\begin{minipage}{\imgsize\textwidth}\includegraphics[width=\textwidth]{images/#1/t2.png}\end{minipage}
\begin{minipage}{\betweenimgsize\textwidth}\subfloat[$l_\text{GS}$]{~}\end{minipage}
\begin{minipage}{\imgsize\textwidth}\includegraphics[width=\textwidth]{images/#1/gtb.png}\end{minipage}

\begin{minipage}{\betweenimgsize\textwidth}\subfloat[$\hat{p}_B^{(i)}$]{~}\end{minipage}
\begin{minipage}{\imgsize\textwidth}\includegraphics[width=\textwidth]{images/#1/flairinp.png}\end{minipage}
\begin{minipage}{\imgsize\textwidth}\includegraphics[width=\textwidth]{images/#1/t1ceinp.png}\end{minipage}
\begin{minipage}{\imgsize\textwidth}\includegraphics[width=\textwidth]{images/#1/t1inp.png}\end{minipage}
\begin{minipage}{\imgsize\textwidth}\includegraphics[width=\textwidth]{images/#1/t2inp.png}\end{minipage}
\begin{minipage}{\betweenimgsize\textwidth}\subfloat[$\hat{l}_\mathcal{B}$]{~}\end{minipage}
\begin{minipage}{\imgsize\textwidth}\includegraphics[width=\textwidth]{images/#1/lm.png}\end{minipage}

\begin{minipage}{\betweenimgsize\textwidth}\subfloat[$\hat{x}_A^{(i)}$]{~}\end{minipage}
\begin{minipage}{\imgsize\textwidth}\subfloat[FLAIR]{\includegraphics[width=\textwidth]{images/#1/resflair.png}}\end{minipage}
\begin{minipage}{\imgsize\textwidth}\subfloat[T1CE]{\includegraphics[width=\textwidth]{images/#1/rest1ce.png}}\end{minipage}
\begin{minipage}{\imgsize\textwidth}\subfloat[T1]{\includegraphics[width=\textwidth]{images/#1/rest1.png}}\end{minipage}
\begin{minipage}{\imgsize\textwidth}\subfloat[T2]{\includegraphics[width=\textwidth]{images/#1/rest2.png}}\end{minipage}
\begin{minipage}{\betweenimgsize\textwidth}\subfloat[$l_\text{M}$]{~}\end{minipage}
\begin{minipage}{\imgsize\textwidth}\subfloat[prob. map]{\includegraphics[width=\textwidth]{images/#1/mdgru.png}}\end{minipage}

}

\newcommand{\mynetworkgraphnohealthyencodingforward}[5]{
\begin{tikzpicture}[auto,  node distance=1.5cm, >=triangle 45,scale=#5, every node/.style={scale=#5}, <-/.style={latex-}, ->/.style={-latex}, edge/.style={-latex}]

  \tikzstyle{smallcircle}=[circle,draw=black!100,fill=black!100,minimum size=0mm,inner sep=1pt]
  \tikzstyle{multcircle}=[circle,cross,draw=black!100,fill=white!100,minimum size=0mm,inner sep=3pt]
  \tikzstyle{encA}=[regular polygon,regular polygon sides=3,shape border rotate=270,draw=blue!75,fill=gray!10,minimum size=0pt,inner sep=0pt]
  \tikzstyle{encB}=[regular polygon,regular polygon sides=3,shape border rotate=270,draw=blue!75,fill=gray!40,minimum size=0pt,inner sep=0pt]
  \tikzstyle{decB}=[regular polygon,regular polygon sides=3,shape border rotate=90,draw=red!75,fill=gray!40,minimum size=0pt,inner sep=0pt]
  \tikzstyle{decA}=[regular polygon,regular polygon sides=4,shape border rotate=90,draw=red!75,fill=gray!10,minimum size=0pt,inner sep=0pt]
    \tikzstyle{disc}=[regular polygon,regular polygon sides=4,shape border rotate=90,draw=black!75,fill=gray!60,minimum size=0pt,inner sep=0pt]
\begin{scope}[local bounding box=inputs]
\draw
	node at (-0.5,1)[disc,name=DisA]{$D_#1$}
 	node at (0,1.45)[name=dB]{}
	node at (0,0)[name=A]{$x_#1$};
	\draw [->,draw=black!50!green!50](A) to [out=135, in=300] node {} (DisA.south);

\end{scope}
\begin{scope}[local bounding box=scope1,shift=(inputs.west),xshift=-10mm,yshift=5mm]
\draw
      node at (0.5,-1) (GammaA) {}
      node at (1.5,-1) (gA) [] {}
       node at (2.75,-0.65) [decA] (G) {$#3$}
       node at (4.5, -1) [smallcircle] (LGhat) {}
       node at (3.8, -1) [smallcircle] (PGhat) {}
       node at (3.8, -1.7) [smallcircle] (gateA) {}
       node at (4,-1.35) [smallcircle] (gateAnode) {}
       node at (4,1) [multcircle] (multcircleDA) {}
       node at (5,1) [encB] (DA) {$\Delta_#1$};

         \draw (A) |- node {} (gateA);
         \draw (G.north east) to [out=0, in=90] node [label={[shift={(0,-0.2)}]$\hat{l}_#1$}] {} (LGhat);
         \draw (G.south east) -| node [label={[shift={(-0.1,-0.1)}]$\hat{p}_#1$}] {} (PGhat);
         \draw (LGhat) -- node {} (gateAnode);
         \draw (G.north east) to [out=0, in=270] node {} (multcircleDA.south);
         \draw (A) to [out=90, in=180] node {} (multcircleDA.west);
         \draw (multcircleDA.east) -- node {} (DA.west);
         \draw [->](A) to [out=90, in=180] node {} (G);
\end{scope}
\begin{scope}[local bounding box=intermediates,shift=(scope1.east |- inputs),xshift=5mm,yshift=-5mm]
\draw 	node at (0,2)[name=dA]{$\hat{\delta}_#1$}
	node at (0,0)[name=Bhat]{$\hat{y}_#2$}
	edge [pre] (LGhat)
	node at (-0.5,1)[disc,name=DisB]{$D_#2$};
\draw [->,draw=red!50](Bhat) to [out=135, in=300] node {} (DisB.south);

\end{scope}
\begin{scope}[local bounding box=scope2,shift=(intermediates.west |- inputs),xshift=23mm,yshift=5mm]
\draw

     node at (0.5,-1) [encB] (GammaB) {$\Gamma_#2$}
       edge [pre,<-] (Bhat)
     node at (1.5,-1) (gB) [] {$\hat{\gamma}_{#2}$}
       node at (2.75,-0.65) [decB] (F) {$#4$}
       node at (5,1) (DB) {}
       node at (4.5, -1) [smallcircle] (LFtilde) {}
       node at (3.8, -1) [smallcircle] (PFtilde) {}
       node at (3.8, -1.7) [smallcircle] (gateB) {}
       node at (4,-1.35) [smallcircle] (gateBnode) {};
       node at (4,1)  (multcircleDB) {};

         \draw [->](gB) to [out=90, in=180] node {} (F.west);
         \draw [->](dA) to [out=0, in=180] node {} (F.west);
         \draw (Bhat) |- node {} (gateB);
         \draw (F.north east) to [out=0, in=90] node [label={[shift={(-0.5,-0.1)}]$\tilde{l}_#2$}] {} (LFtilde);
         \draw (F.south east) -| node [label={[shift={(-0.1,-0.1)}]$\tilde{p}_#2$}] {} (PFtilde);
         \draw (LFtilde) -- node {} (gateBnode);

\end{scope}
\begin{scope}[local bounding box=intermediates,shift=(scope2.east |- inputs),xshift=5mm,yshift=-5mm]
\draw 	
	node at (0,0)[name=Atilde]{$\tilde{x}_#1$}
	edge [<-,pre] (LFtilde);
\end{scope}
\end{tikzpicture}
}

\newcommand{\mynetworkgraphnohealthyencodingbackward}[5]{
\begin{tikzpicture}[auto, node distance=1.5cm, >=triangle 45,scale=#5, every node/.style={scale=#5},<-/.style={latex-}, ->/.style={-latex}, edge/.style={-latex}]
  \tikzstyle{smallcircle}=[circle,draw=black!100,fill=black!100,minimum size=0mm,inner sep=1pt]
  \tikzstyle{multcircle}=[circle,cross,draw=black!100,fill=white!100,minimum size=0mm,inner sep=3pt]
  \tikzstyle{encA}=[regular polygon,regular polygon sides=3,shape border rotate=270,draw=blue!75,fill=gray!10,minimum size=0pt,inner sep=0pt]
  \tikzstyle{encB}=[regular polygon,regular polygon sides=3,shape border rotate=270,draw=blue!75,fill=gray!40,minimum size=0pt,inner sep=0pt]
  \tikzstyle{decA}=[regular polygon,regular polygon sides=3,shape border rotate=90,draw=red!75,fill=gray!40,minimum size=0pt,inner sep=0pt]
  \tikzstyle{decB}=[regular polygon,regular polygon sides=4,shape border rotate=90,draw=red!75,fill=gray!10,minimum size=0pt,inner sep=0pt]
    \tikzstyle{disc}=[regular polygon,regular polygon sides=4,shape border rotate=90,draw=black!75,fill=gray!60,minimum size=0pt,inner sep=0pt]

\begin{scope}[local bounding box=inputs,anchor=west,xshift=100mm]

\draw 	node at (-0.5,1)[disc,name=DisA]{$D_#1$}
	node at (0,2)[name=dB]{$\delta_#2$}
	node at (0,0)[name=A]{$x_#1$};
\draw [->,draw=black!50!green!50](A) to [out=135, in=300] node {} (DisA.south);

\end{scope}
\begin{scope}[local bounding box=scope1,shift=(inputs.west),xshift=15mm,yshift=0mm]
\draw

     node at (0.5,-1) [encB] (GammaA) {$\Gamma_#1$}
       edge [pre] (A)
     node at (1.5,-1) (gA) [] {$\hat{\gamma}_{#1}$}
       node at (2.75,-0.65) [decA] (G) {$#3$}
       node at (4.5, -1) [smallcircle] (LGhat) {}
       node at (3.8, -1) [smallcircle] (PGhat) {}
       node at (3.8, -1.7) [smallcircle] (gateA) {}
       node at (4,-1.35) [smallcircle] (gateAnode) {};
       node at (4,1)  (multcircleDA) {};

         \draw [->](gA) to [out=90, in=180] node {} (G.west);
         \draw [->](dB) to [out=0, in=180] node {} (G.west);
         \draw (A) |- node {} (gateA);
         \draw (G.north east) to [out=0, in=90] node [label={[shift={(-0.5,-0.1)}]$\hat{l}_#1$}] {} (LGhat);
         \draw (G.south east) -| node [label={[shift={(-0.1,-0.1)}]$\hat{p}_#1$}] {} (PGhat);
         \draw (LGhat) -- node {} (gateAnode);

\end{scope}
\begin{scope}[local bounding box=intermediates,shift=(scope1.east |- inputs),xshift=10mm,yshift=-10mm]
\draw 	
	node at (0,0)[name=Bhat]{$\hat{y}_#2$}
	edge [pre] (LGhat)
	node at (-0.5,1)[disc,name=DisB]{$D_#2$};
	\draw [->,draw=red!50](Bhat) to [out=135, in=300] node {} (DisB.south);

\end{scope}
\begin{scope}[local bounding box=scope2,shift=(intermediates.west |- inputs),xshift=-8mm,yshift=0mm]
\draw
      node at (0.5,-1) (GammaA) {}
      node at (1.5,-1) (gA) [] {}
       node at (2.75,-0.65) [decB] (F) {$#4$}
       node at (4.5, -1) [smallcircle] (LFtilde) {}
       node at (3.8, -1) [smallcircle] (PFtilde) {}
       node at (3.8, -1.7) [smallcircle] (gateA) {}
       node at (4,-1.35) [smallcircle] (gateAnode) {}
       node at (4,1) [multcircle] (multcircleDA) {}
       node at (5,1) [encB] (DA) {$\Delta_#2$};

         \draw (Bhat) |- node {} (gateA);
         \draw (F.north east) to [out=0, in=90] node [label={[shift={(0,-0.2)}]$\tilde{l}_#2$}] {} (LFtilde);
         \draw (F.south east) -| node [label={[shift={(-0.1,-0.1)}]$\tilde{p}_#2$}] {} (PFtilde);
         \draw (LFtilde) -- node {} (gateAnode);
         \draw (F.north east) to [out=0, in=270] node {} (multcircleDA.south);
         \draw (Bhat) to [out=90, in=180] node {} (multcircleDA.west);
         \draw (multcircleDA.east) -- node {} (DA.west);
         \draw [->](Bhat) to [out=90, in=180] node {} (F);
\end{scope}
\begin{scope}[local bounding box=intermediates,shift=(scope2.east |- inputs),xshift=5mm,yshift=-10mm]
\draw 	node at (0,2)[name=dAtilde]{$\tilde{\delta}_#2$}
	node at (0,0)[name=Atilde]{$\tilde{x}_#1$}
	edge [pre] (LFtilde);
\end{scope}
\end{tikzpicture}
}
\title{Pathology Segmentation using Distributional Differences to Images of Healthy Origin}
\author{Simon Andermatt \and Antal Horv\'{a}th \and Simon Pezold \and Philippe Cattin}

\institute{Department of Biomedical Engineering, 
University of Basel,
Allschwil,
Switzerland
}

\begin{document}
\tikzset{
    -|/.style={to path={-| (\tikztotarget)}},
    |-/.style={to path={|- (\tikztotarget)}},
      block/.style    = {draw, thick, rectangle, minimum height = 3em,
    minimum width = 3em},
  sum/.style      = {draw, circle, node distance = 2cm},
  input/.style    = {coordinate},
  output/.style   = {coordinate},
  cross/.style={path picture={ 
  \draw[black]
(path picture bounding box.south east) -- (path picture bounding box.north west) (path picture bounding box.south west) -- (path picture bounding box.north east);
}}
}
\maketitle
\begin{abstract}
Fully supervised segmentation methods require a large training cohort of already segmented images, providing information at the pixel level of each image. We present a method to automatically segment and model pathologies in medical images, trained solely on data labelled on the image level as either healthy or containing a visual defect. We base our method on CycleGAN, an image-to-image translation technique, to translate images between the domains of healthy and pathological images. We extend the core idea with two key contributions. Implementing the generators as residual generators allows us to explicitly model the segmentation of the pathology. Realizing the translation from the healthy to the pathological domain using a variational autoencoder allows us to specify one representation of the pathology, as this transformation is otherwise not unique. Our model hence not only allows us to create pixelwise semantic segmentations, it is also able to create inpaintings for the segmentations to render the pathological image healthy. Furthermore, we can draw new unseen pathology samples from this model based on the distribution in the data. We show quantitatively, that our method is able to segment pathologies with a surprising accuracy being only slightly inferior to a state-of-the-art fully supervised method, although the latter has per-pixel rather than per-image training information. Moreover, we show qualitative results of both the segmentations and inpaintings. Our findings motivate further research into weakly-supervised segmentation using image level annotations, allowing for faster and cheaper acquisition of training data without a large sacrifice in segmentation accuracy.
\end{abstract}
\section{Introduction}
Supervised segmentation in medical image analysis is an almost solved problem for many applications, where methodological improvements have a marginal effect on accuracy.
Such methods depend on a large annotated training corpus, where pixelwise labels have to be provided by medical experts. In practice, such data are expensive to gather. In contrast, weakly labelled data, such as images showing a certain disease are easily obtainable, since they are created on a daily basis in medical practice. We want to take advantage of these data for pathology segmentation, by providing a means to finding the difference between healthy and pathological data distributions. We present a weakly supervised framework capable of pixelwise segmentation as well as generating samples from the pathology distribution, trained on images which are only annotated with a scalar binary label at the image level classifying them as healthy or pathological. 

Our idea is inspired by CycleGAN \cite{cyclegan}, a recently proposed solution for unpaired image to image translation, where the combination of domain-specific generative adversarial networks (GANs) and so-called cycle consistency allow for robust translation. We call our adaptation PathoGAN and count the following contributions: We formulate a model capable of segmentation based on a single label per training sample. We simultaneously train two generative models, able to generate inpaintings at a localized region of interest to transform an image from one domain to the other. We are able to sample healthy anatomy as well as sample possible pathologies for a given anatomical structure. Furthermore, our method enforces domain-specific information to be encoded outside of the image, which omits adversarial ``noise'' common to CycleGAN \cite{steganography} to some degree. 

We show the performance of our implementation on 2d slices of the training data of the Brain Tumor Segmentation Challenge 2017 \cite{brats2015short,brats2017} and compare our segmentation performance to a supervised segmentation technique \cite{mdgru}.

CycleGAN has been previously used to segment by transfering to another target modality, where segmentation maps are available (e.g. \cite{notargetmodality2}), or applied to generate training from cheaply generated synthetic labelmaps \cite{gentraining}. Using a Wasserstein GAN, another method directly estimates an additive visual attribution map \cite{visualfeatureattribution}. To our knowledge, there has not been a method that jointly learns to segment on one medical imaging modality using only image-level labels and generate new data using GANs for both healthy and pathological cases.
\section{Methods}
\subsection{Problem Statement}
We assume two image domains $\mathcal{A}$ and $\mathcal{B}$, where the former contains only images of healthy subjects and the latter consists of images showing a specific pathology. We seek to approximate the functions $G_\mathcal{A}$ and $G_\mathcal{B}$ that perform the mappings $(x_\mathcal{A},\delta_\mathcal{B}) \mapsto \hat{y}_{B}$ and $(x_\mathcal{B},\delta_\mathcal{A})  \mapsto \hat{y}_{\mathcal A}$, where $x_\mathcal{A}, \hat{y}_\mathcal{A} \in \mathcal{A}$ and $ x_\mathcal{B}, \hat{y}_{\mathcal B} \in \mathcal{B}$. Vectors $\delta_\mathcal{B}$ and $\delta_\mathcal{A}$ encode the missing target image information (e.g. the pathology): 
\begin{align}
 \label{eq:weighting}\hat{y}_\mathcal{B} = G_\mathcal{A}(x_\mathcal{A}, \delta_\mathcal{B})
 ,\hspace{1em} \hat{y}_\mathcal{A} = G_\mathcal{B}(x_\mathcal{B}, \delta_\mathcal{A}).
\end{align}
We encourage $G_\mathcal{A}, G_\mathcal{B}$ to produce results, such that the mappings are realistic (\ref{eq:believable}), cycle-consistent (\ref{eq:reversible}), specific (\ref{eq:specific}) and that only the affected part in the image is modified (\ref{eq:minimal}):

\begin{alignat}{3}
 G_{\mathcal{A}}(x_{\mathcal{A}}, \delta_{\mathcal{B}}) &\sim \mathcal{B}, & G_{\mathcal{B}}(x_{\mathcal{B}}, \delta_{\mathcal{A}}) & \sim \mathcal{A}\label{eq:believable},\\
 G_\mathcal B(G_\mathcal A(x_\mathcal A, \delta_\mathcal B), \delta_\mathcal A) &\approx x_\mathcal A, &G_\mathcal A(G_\mathcal B(x_\mathcal B, \delta_\mathcal A), \delta_\mathcal B) & \approx x_\mathcal B, \label{eq:reversible}\\
 G_\mathcal B(x_\mathcal A,0) &\approx x_\mathcal A, & G_\mathcal A(x_\mathcal B, 0) & \approx x_\mathcal B,\label{eq:specific}\\
 \argmin\limits_{\overbar{G}_{\mathcal A}} |x_{\mathcal A}-\overbar{G}_{\mathcal A}(x_\mathcal A,\delta_\mathcal B)| &\approx G_\mathcal A, \hspace{3em}&\argmin\limits_{\overbar{G}_{\mathcal B}} |x_{\mathcal B}-\overbar{G}_{\mathcal B}(x_\mathcal B, \delta_\mathcal A)| & \approx G_\mathcal B.\label{eq:minimal}
\end{alignat}
\subsection{Model Topology}
To fulfill Eqs. (\ref{eq:believable}--\ref{eq:minimal}), we adopt the main setup and objective from CycleGAN: we employ two discriminators, $D_\mathcal{A}$ and $D_\mathcal{B}$ together with generators $G_\mathcal{A}$ and $G_\mathcal{B}$ to perform the translation from domain $\mathcal{A}$ to $\mathcal{B}$ and vice versa, formulating two generative adversarial networks (GANs) \cite{gan}. In both directions, the respective discriminator is trained to distinguish a real image from the output of the generator, whereas the generator is incentivized to generate samples that fool the discriminator. 

In the remaining paper, we will use a short notation for equations which are applicable to both pathways (see Fig. \ref{simplearchitecture}) to overcome redundancy due to symmetrical components. We specify placeholder domains $\mathcal X$ and $\mathcal Y$, where for both pathways, $\mathcal X$ denotes the domain the initial image belongs to and $\mathcal Y$ stands for the target domain we want to transfer an image into. Domain $\mathcal X$ could therefore be either $\mathcal A$ or $\mathcal B$, and domain $\mathcal Y$ either $\mathcal B$ or $\mathcal A$, respectively. Furthermore, $x_\mathcal X \in \mathcal X$ denotes the image $x$ in the respective initial domain $\mathcal X$, $G_\mathcal X(x_\mathcal X) = \hat{y}_\mathcal Y$ its transformation in the target domain $\mathcal Y$ and $G_\mathcal Y(\hat{y}_\mathcal Y) = \tilde{x}_\mathcal X$ its reconstruction in $\mathcal X$. 

\subsubsection{Residual Generator}
\begin{figure}
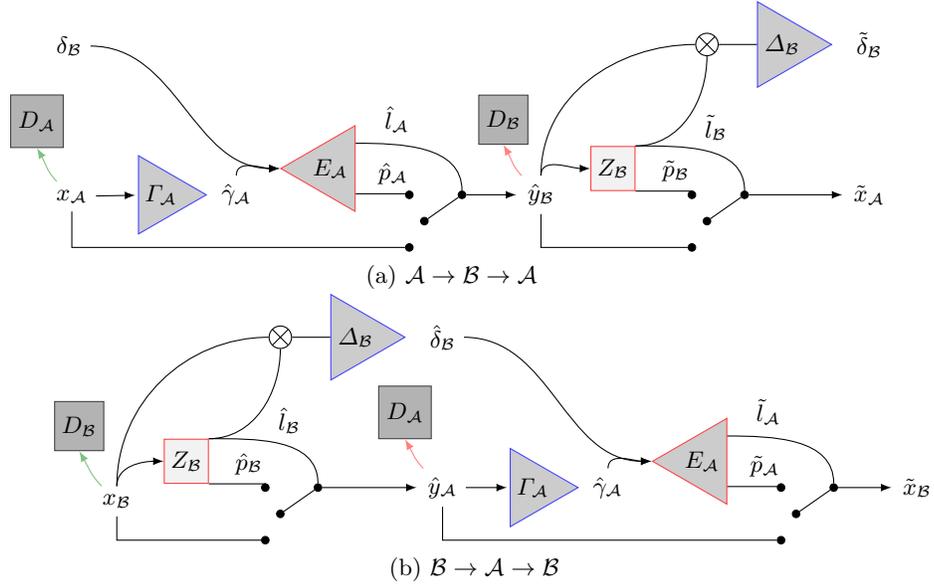

\newcommand{\suma}{\Large$+$}
\newcommand{\inte}{$\displaystyle \int$}
\newcommand{\derv}{\huge$\frac{d}{dt}$}
\def\scale{1}
\begin{minipage}{\textwidth}\centering
\subfloat[$\mathcal{A}\rightarrow\mathcal{B}\rightarrow\mathcal{A}$]{
  \mynetworkgraphnohealthyencodingbackward{\mathcal{A}}{\mathcal{B}}{{E_\mathcal{A}}}{{Z_\mathcal{B}}}{\scale}
}
\end{minipage}
 \begin{minipage}{\textwidth}\centering
 \subfloat[$\mathcal{B}\rightarrow\mathcal{A}\rightarrow\mathcal{B}$]{
   \mynetworkgraphnohealthyencodingforward{\mathcal{B}}{\mathcal{A}}{{Z_\mathcal{B}}}{{E_\mathcal{A}}}{\scale}
 }\end{minipage}
\caption{Proposed architecture: \emph{top to bottom:} directions $\mathcal{A}\rightarrow\mathcal{B}\rightarrow\mathcal{A}$ and $\mathcal{B}\rightarrow\mathcal{A}\rightarrow\mathcal{B}$; $x_\mathcal{A}, x_\mathcal{B}$ are samples from the two data distributions $\mathcal{A}$ (healthy), $\mathcal{B}$ (pathological) and $\delta_\mathcal{B} \sim \mathcal{N}(0,I)$. Red and blue triangles depict decoder and encoder networks. A red square illustrates a simple generator. $\Delta_\mathcal{B}$ and $\Gamma_\mathcal{A}$ encode features inside and outside the pathological region. $\Delta_\mathcal{B}, \Gamma_\mathcal{A}$ and $E_\mathcal{A}$ form a variational autoencoder, which acts as the residual generator $Z_\mathcal A$ used in $G_\mathcal A$. On the other hand, the residual generator $Z_\mathcal B$ in $G_\mathcal B$ is implemented as one component, as information about the missing healthy structure is completely inferred from the surroundings without explicitly encoding it.}
\label{simplearchitecture}
\end{figure}

In order to segment pathologies, we seek to only modify a certain part of the image. In contrast to CycleGAN, we model the transformation $G$ from one domain to the other as a residual or inpainting $p$ which is exchanged with part $l$ of the original image. We achieve this by directly estimating $n + 1$ feature maps $r_\mathcal X$ using a residual generator $Z_\mathcal X$ within $G_\mathcal X$, where $n$ is the number of image channels used. We obtain labelmap $l_\mathcal X$ and inpaintings $p_\mathcal X$, activating $r_\mathcal X^{(0)}$ with a sigmoid and each $r_\mathcal X^{(i)}$ with a $\tanh$ activation:
\begin{align}
l_\mathcal X = \Sigmoid{\left(r_\mathcal X^{(0)} + \epsilon \right)}, \hspace{1em}p_\mathcal X^{(i-1)} = \tanh{\left(r_\mathcal X^{(i)}\right)}, \hspace{1em} r_\mathcal X = Z_\mathcal X (x_\mathcal X, \delta_\mathcal Y),
\end{align}
where $\Sigmoid{(y) = \frac{1}{1+e^{-y}}}$ and $i > 0$. With $\epsilon \sim \mathcal{N}(0,I)$, we turn $r_\mathcal X^{(0)} + \epsilon$ into samples from $\mathcal{N}(r_\mathcal X^{(0)},I)$ using the reparameterization trick \cite{variationalautoencoder}. This allows reliable calculations of $l_\mathcal X$ only for large absolute values of $r_\mathcal X^{(0)}$, forcing $l_\mathcal X$ to be binary and intensity information to be encoded in the inpaintings. We set $\epsilon$ to zero during testing.
From $l_\mathcal X$ and $p_\mathcal X$ we compute the translated result $\hat{y}_\mathcal Y$, supposedly in domain $\mathcal Y$ now:
\[\hat{y}_\mathcal Y = l_\mathcal X \odot p_\mathcal X + (1 - l_\mathcal X) \odot x_\mathcal X = G_\mathcal{X}(x_\mathcal X, \delta_\mathcal Y).\]
In the following, we detail the computation of $r_\mathcal{A}$ and $r_\mathcal{B}$ using the two networks $Z_\mathcal A$ and $Z_\mathcal B$ for the two possible translation directions. Both translation pathways are visualized in Fig. \ref{simplearchitecture}.
\subsubsection{$Z_\mathcal{A}: \mathcal{A} \rightarrow \mathcal{B}$}
To map from healthy to pathological data, we estimate $r_\mathcal{A}$ (and thus $l_\mathcal{A},p_\mathcal{A}$) using a variational autoencoder (VAE) \cite{variationalautoencoder}. First, we employ encoders $\Gamma_\mathcal{A}$ and $\Delta_\mathcal{B}$ to encode anatomical information around and inside the pathological region: 
\[\gamma_{\mathcal{A}} = \Gamma_{\mathcal A}(x_\mathcal{A}), \hspace{3em} \delta_{\mathcal{B}} = \Delta_{\mathcal B}(l_\mathcal{B}' \odot {x_{\mathcal{B}}'}), \]
where $x_\mathcal{A}$ is our healthy image, $l_\mathcal{B}'$ and $x_\mathcal{B}'$ are the labelmap and pathological image of the previous transformation and $\delta_\mathcal{B}, \gamma_\mathcal{A} \sim \mathcal{N}(0,I)$. If $l_\mathcal{B}'$ and $x_\mathcal{B}'$ are not available because $x_\mathcal{A}$ is a real healthy image, we simply sample $\delta_\mathcal B$. 
Finally, a decoder $E_\mathcal{A}$ is applied to $\gamma_\mathcal{A}$ and $\delta_\mathcal{B}$:
\[r_\mathcal{A} = E_\mathcal{A}(\gamma_{\mathcal{A}}, \delta_\mathcal{B}).\]
The residual generator $Z_\mathcal{A}$ is hence composed of encoder $\Gamma_\mathcal{A}$ and decoder $E_\mathcal A$, where $\delta_\mathcal B$ is either sampled or calculated with the additional encoder $\Delta_\mathcal{B}$:
\[Z_\mathcal{A}(x_\mathcal A, \delta_\mathcal B) = E_\mathcal{A}(\Gamma_\mathcal B(x_\mathcal A), \delta_\mathcal B),\]
\subsubsection{$Z_\mathcal{B}: \mathcal{B} \rightarrow \mathcal{A}$}
To generate healthy samples from pathological images, we use the residual generator $Z_\mathcal{B}$ directly on the input as introduced in \cite{cyclegan} and estimate $r$ directly:
\[r_\mathcal{B} = Z_\mathcal{B}(x_\mathcal B).\]
Here, we omit $\delta_\mathcal{A}$, since the location and appearance of missing healthy tissue can be inferred from $x_\mathcal B$. We also omit using an encoding bottleneck due to possible information loss and less accurate segmentation. 

\subsection{Objective}
To train this model, a number of different loss terms are necessary. In the following, we explain the individual components using $\hat{y}_.$ and $\tilde{x}_.$ to denote results from the translated and reconstructed images respectively (e.g. mapping $x_\mathcal X$ into $\mathcal Y$ results in $\hat{y}_\mathcal Y$, translating it back results in $\tilde{x}_\mathcal X$). We use $\lambda_.$ to weight the contribution of different loss terms.

\subsubsection{CycleGAN}
As in CycleGAN \cite{cyclegan}, we formulate a least squares proxy GAN loss, which we minimize with respect to $G_\mathcal X$ and maximize with respect to $D_\mathcal Y$:
\begin{align}
\mathcal{L}_{\text{GAN}}(D_\mathcal Y, G_\mathcal X, x_\mathcal X, x_\mathcal Y) &=  \mathbb{E}[(D_\mathcal Y(x_\mathcal Y))^2] + \mathbb{E}[(1-D_\mathcal Y(G_\mathcal X(x_\mathcal X,\delta_\mathcal Y)))^2].
\end{align}
Likewise, to make mappings reversible, we add the cycle-consistency loss:
\begin{align}
\mathcal{L}_{\text{CC}}(G_\mathcal X, G_\mathcal Y, x_\mathcal X) &= \lambda_{\text{CC}}||G_\mathcal Y(G_\mathcal X(x_\mathcal X, \delta_\mathcal Y),\delta_\mathcal X)-x_\mathcal X||_1.
\end{align}
$\mathcal{L}_{\text{GAN}}$ and $\mathcal{L}_{\text{CC}}$ encourage the properties defined in Eqs.~(\ref{eq:believable}) and (\ref{eq:reversible}).

\subsubsection{Variational Autoencoder}
A variational autoencoder (VAE) is trained by minimizing the KL-divergence of the distribution $q(z|x)$ of encoding $z$ to some assumed distribution and the expected reconstruction error $\log p (x | z)$, where $x$ is the data. In contrast to a classical VAE, we use two distinct encoding vectors $\gamma_\mathcal{A}$ and $\delta_\mathcal{B}$, encoding the healthy and the pathological part, and produce two separate results, the labelmap $l_\mathcal{A}$ and the inpainting $p_\mathcal{A}$. We directly calculate the KL-divergence for our two encodings:
\begin{align}
 \mathcal{L}_{\text{KL}}(G_\mathcal{A}, G_\mathcal{B}, x_\mathcal{A},x_\mathcal{B})=\text{KL}[q(\gamma_\mathcal{A}|x_\mathcal{A})||\mathcal{N}(0,I)]+\text{KL}[q(\delta_\mathcal{B}|x_\mathcal{B}, \hat{l}_\mathcal{B})||\mathcal{N}(0,I)].
\end{align}
For the expected reconstruction error, we assume that $l$ and $p$ follow approximately a Bernoulli and a Gaussian distribution ($\mathcal{N}(\mu,I)$). We selectively penalize the responsible encoding, by using separate loss functions for the residual region and the rest. Unfortunately, we only ever have access to the ground truth of one of these regions, since we do not use paired data. We solve this, by using the relevant application in the network, where individual ground truths are available, to calculate the approximation of the marginal likelihood lower bound:
\begin{align}
\begin{split}
&\mathcal{L}_{\text{VAE}}(G_\mathcal X, G_\mathcal Y, x_\mathcal X, x_\mathcal Y) = \\&- \frac{\lambda_{\text{VAE}}}{N}\sum\limits_{m=1}^N (\log p(\tilde{l}_\mathcal Y|\gamma_\mathcal X, \delta_\mathcal Y) + \log p(\hat{p}_\mathcal X|\gamma_\mathcal X) + \log p(\tilde{p}_\mathcal Y|\delta_\mathcal Y)),
\end{split}
\end{align}
where $\hat{p}_\mathcal X$ denotes the inpainting used to translate the original image $x_\mathcal X$ to domain $\mathcal Y$ and $N$ is the total number of pixels. $\tilde{p}_\mathcal X$ is the inpainting produced when translating an already translated image $\hat{y}_\mathcal X$ that originated from $\mathcal Y$ back to that domain. Similarly, $\hat{l}_\mathcal X$ and $\tilde{l}_\mathcal X$ denote the respective labelmaps:
\begin{align}
\log p(\hat{p}_\mathcal X|\gamma_\mathcal X) &= \frac{||(1-\hat{l}_\mathcal X)(\hat{p}_\mathcal X-x_\mathcal X)||_2}{\omega_1},\\
\log p(\tilde{p}_\mathcal X|\delta_\mathcal Y) &= \frac{||\tilde{l}_\mathcal Y(\tilde{p}_\mathcal X-x_\mathcal Y)||_2}{\omega_2},
\end{align}
where $\omega_1 = \frac{\sum(1-\hat{l}_\mathcal X)+\varepsilon}{N}$ and $\omega_2 = \frac{\sum(\tilde{l}_\mathcal Y)+\varepsilon}{N}$ are considered constant during optimization, with $\varepsilon>0$. 
Finally, we use the labelmap produced by the other generator responsible for the opposite transformation $\hat{l}_\mathcal Y$ as ground truth for $\tilde{l}_\mathcal X$, where we consider $\hat{l}_\mathcal Y$ constant in this term:
\begin{align}
 \log p(\tilde{l}_\mathcal X|\gamma_\mathcal X, \delta_\mathcal Y) =  \hat{l}_\mathcal Y \log \tilde{l}_\mathcal X + (1-\hat{l}_\mathcal Y) \log (1-\tilde{l}_\mathcal X).
\end{align}
To restrict the solution space of our model, we use $\mathcal{L}_\text{VAE}$ for both directions.
\subsubsection{Identity Loss}
We apply an identity loss \cite{cyclegan} on labelmap $l_{\mathcal X,x_\mathcal Y}$ which results from feeding $G_\mathcal X$ with the wrong input $x_\mathcal Y$. In this case $G_\mathcal X$ should not change anything, since the input is already in the desired domain $\mathcal Y$:
\begin{align}
\mathcal{L}_{\text{Idt}}(G_\mathcal X, x_\mathcal Y) = \lambda_{\text{Idt}}||l_{\mathcal X,x_\mathcal Y}||_1. 
\end{align}
\subsubsection{Relevancy Loss}
By now, we have defined all necessary constraints for a successful translation between image domains. The remaining constraints restrict the location and amount of change, $l_\mathcal X$. Fulfilling \eqref{eq:minimal}, we want to entice label map $l_\mathcal X$ to be only set at locations of a large difference between inpainting $p_\mathcal X$ and image $x_\mathcal X$ and penalize large label maps in general:
\begin{align}
\mathcal{L}_{\text{R}}(G_\mathcal X,x_\mathcal X) = \lambda_{\text{R}}\left[||-\log(1-l_\mathcal X^2)||_1-\frac{|| l_\mathcal X (x_\mathcal X-p_\mathcal X)||_1}{|| l_\mathcal X||_1}\right].
\end{align}
In order to not reward exaggerated pathology inpaintings, we consider $(x_\mathcal X-p_\mathcal X)$ constant in this expression.
\subsubsection{Full PathoGAN Objective}
combining all loss terms for direction $\mathcal X$ to $\mathcal Y$ as $\mathcal{L}_{\mathcal X\rightarrow \mathcal Y}$, we can finally define:
\begin{align}
\mathcal{L}_{\mathcal X\rightarrow \mathcal Y} &= \mathcal{L}_{\text{GAN}} + \mathcal{L}_{\text{CC}} + \mathcal{L}_{\text{VAE}} + \mathcal{L}_{\text{Idt}} + \mathcal{L}_{\text{R}},\\
\mathcal{L}_{\text{PathoGAN}} &= \mathcal{L}_{{\mathcal{A}}\rightarrow {\mathcal{B}}} +\mathcal{L}_{{\mathcal{B}}\rightarrow{\mathcal{A}}}+\lambda_{\text{KL}}  \mathcal{L}_{\text{KL}}(x_\mathcal{A},x_\mathcal{B}).
\end{align}

\subsection{Network structure}

We describe the networks used in our generators $G_\mathcal A$ and $G_\mathcal B$ using a shorthand notation which we will define in the following. We denote the number of output feature maps as \lstinline {f}, the number of inpaintings including labelmap as \lstinline {r}, the smallest image width and height before reshaping to a vector as \lstinline{i} and the encoding size as \lstinline{z}. The smallest image width and height in our case is $15$ and we chose an encoding length of $256$. We denote the reshaping operation from a square image to a vector as \lstinline{Q2F} and the inverse operation as \lstinline{F2Q}. We further denote a convolution operation with stride $1$ and a kernel size of \lstinline{k} as \lstinline{Ck-f}. We use a lower-case \lstinline{c} in \lstinline{ck-f} if the convolution is followed by an instance norm and an activation with an exponential linear unit (ELU) \cite{elu}. A downsampling operation of a stride $2$ convolution with kernel size $3$, followed with an instance norm and following ELU is coded as \lstinline{df}, whereas an upsampling operation using a transposed convolution followed by an instance norm and an ELU activation is denoted as \lstinline{uf}. We describe a fully connected layer with \lstinline{l(f)} and a residual block as \lstinline{Rf}. A residual block results in a sum of its input with a residual computation of a convolution layer of kernel size $3$, an instance norm, an ELU and a convolution layer of kernel size $3$. Finally, a trailing \lstinline{t} or \lstinline{e} denotes an additional $\tanh$ or ELU activation, respectively.

The generator $G_\mathcal A$ consists of the residual generator $Z_\mathcal A$, which is composed of two encoder networks ($\Gamma_\mathcal A, \Delta_\mathcal B$) and a decoder network ($E_\mathcal A$). Both $\Gamma_\mathcal A$ and $\Delta_\mathcal B$ are defined as:
\begin{lstlisting}
c7-64,d128,d256,d512,d1024,C1-15,Q2F,l(z*i)t,l(2*z).
\end{lstlisting}
The output of these encoders is split in half to produce the mean and log variance vectors of length $256$ each.

Similarly, the decoder network $E_\mathcal A$ is defined as:
\begin{lstlisting}
l(i*i)e,l(i*i),F2Q,c3-1024,u512,u256,C7-256e,R256,R256,R256,
R256,R256,R256,R256,R256,R256,u128,u64,C7-r.
\end{lstlisting}

The generator $G_\mathcal B$ consists of the residual generator network $Z_\mathcal B$, which is slightly adjusted compared to the generators used in \cite{cyclegan}. Here, we use exponential linear units instead of rectified linear units and double each layers' number of feature maps:
\begin{lstlisting}
c7-64,d128,d256,R256,R256,R256,R256,R256,R256,R256,R256,R256,u128,u64,C7-r.
\end{lstlisting}

We use the same network architecture as in CycleGAN \cite{cyclegan} for the discriminator networks. 

\subsection{Data}
We include all training patients of Brats2017 and normalize each brain scan for its non-zero voxels to follow $\mathcal{N}(0,\nicefrac{1}{3})$, and clip the resulting intensities to $[-1,1]$. We select all transverse slices from 60 to 100 in caudocranial direction. In order to create two distinct datasets and relying on the manual segmentations, we label slices without pathology as \emph{healthy}, with more than 20 pixels segmented as \emph{pathological} slices, and discard the rest. For training, we select 1\,500 unaffected and 6\,000 pathological slices from a total of 1\,755 and 9\,413 respectively\footnote{Thus we would like to stress that the manual segmentations were only used to create the two image domains, but not for the actual training.}.

We augment our training data by randomly mirroring the samples, applying a random rotation sampled from $U[-0.1,0.1]$ and a random scaling $s=1.1^r$ where $r$ is sampled from $U[-1,1]$. Furthermore we apply random deformations as described in \cite{mdgru}, with a grid spacing of 128 and sample both components of the grid deformation vectors from $\mathcal{N}(0,5)$.

\begin{table}
{%
\newcommand{\mc}[3]{\multicolumn{#1}{#2}{#3}}
\caption{Segmentation Results. \emph{Columns:} Dice, 95th percentile Hausdorff distance (HD95), average Hausdorff distance (AVD) and volumetric Dice per-patient (Dice PP) by stacking all evaluated slices. \emph{Rows:} Scores are shown as mean$\pm$std(median) for the weakly-supervised PathoGAN (proposed) and the fully-supervised MDGRU, applied to training (Tr) and testing (Te) data.}
\begin{center}
\begin{tabular}{lllllll}
 && Dice (in \%) & HD95 (in pixel)& AVD (in pixel) && Dice PP (in \%)\\
 \midrule
PathoGAN, (Tr) && $72.4\pm24.4(81.0)$&$40.6\pm30.7(38.0)$ &$10.3\pm15.4(4.7)$&&$77.4\pm14.4(81.2)$\\
PathoGAN, (Te) && $72.9\pm23.8(81.4)$&$39.4\pm29.9(37.6)$&$ 9.4\pm13.7(4.6)$&&$77.4\pm14.4(81.7)$\\
MDGRU, (Tr)&& $87.8\pm20.0(94.4)$& $3.7\pm9.7(1.0)$&$1.0\pm4.7(0.2)$&& $90.8\pm8.8(93.3)$\\
MDGRU, (Te)&& $86.3\pm21.3(93.6) $ &$3.9\pm9.5(1.0) $ & $1.1\pm4.9(0.2) $&&$90.6\pm9.5(93.1)$
\end{tabular}
\end{center}
\label{table:quant}
}%
\end{table}
\section{Results}
Since the BratS evaluation is volumetric and comparing performance is difficult, we also train a supervised segmentation technique on our data. We chose MDGRU \cite{mdgru} for this task, a multi-dimensional recurrent neural network, due to code availability and consistent state-of-the-art performance on different datasets.

\begin{figure}
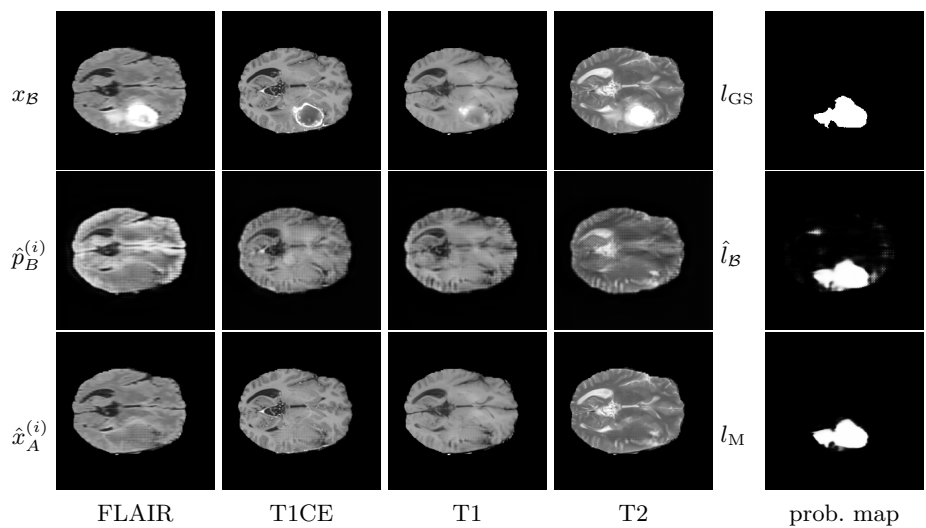

\centering
\minipagetwolayout{HGG-Brats17_2013_10_1-71}
\caption{Qualitative Results of one example from the testing data. \emph{Left columns, top to bottom:} the four available image channels $x_\mathcal B^{(i)}$, the generated inpaintings $\hat{p}_\mathcal B^{(i)}$ and the translated images $\hat{y}_\mathcal A^{(i)}$. \emph{Right column, top to bottom:} The manual segmentation $l_\text{GS}$, the probability maps $\hat{l}_\mathcal B$ from PathoGAN (weakly-supervised, proposed) and $l_\text{M}$ from MDGRU (fully-supervised) for whole tumor.}
\label{qualiresults}
\end{figure}

We trained PathoGAN\footnote{Our implementation is based on \url{https://github.com/junyanz/pytorch-CycleGAN-and-pix2pix}.} for 119 epochs using batches of 4 and $\lambda_{\text{KL}} = 0.1, \lambda_{\text{R}} = 0.5,\lambda_{\text{Idt}} = 1, \lambda_{\text{CC}} = 5$ and $\lambda_{\text{VAE}} = 1$. We trained MDGRU\footnote{We use the implementation of MDGRU at \url{https://github.com/zubata88/mdgru}.} as defined in \cite{mdgru}, using batches of 4 and 27\,500  iterations. Table \ref{table:quant} shows the results on the pathological training and test data. Figure \ref{qualiresults} shows an exemplary sample from the test data. On the left, the input data, the generated inpaintings and the translation result are displayed. On the right, the manual segmentation for ``whole tumor'' and generated labelmaps of the weakly-supervised PathoGAN and fully-supervised MDGRU are presented. 
\section{Discussion}
The results in Fig. \ref{qualiresults} indicate that our relative weighting of the two inpainting reconstruction losses results in better reconstruction inside the tumor region than outside. The labelmaps of the supervised method compared to ours in Fig. \ref{qualiresults} show great agreement, and both are relatively close to the gold standard. As the 95th-percentile and average Hausdorff measures in Table \ref{table:quant} show, there are some outliers in our proposed method, due to its weakly-supervised nature. 
Compared to the fully-supervised method, Dice scores for PathoGAN are about 10\% smaller for both the per-slice and the per-patient case. Slightly inferior results for the proposed method are not surprising, though, given the drastically reduced information density during its training (one label per image for PathoGAN, one label per pixel for fully-supervised MDGRU). Also, it is important to remember that we segment with the only criterion of being not part of the healthy distribution, which could vary from the subjective measures used to manually segment data. The increase in accuracy and decrease in standard deviation in the per-patient case for both methods is most likely caused by the inferior segmentation performance in slices showing little pathology. The per-patient Dice of the supervised method is in the range of the top methods of BraTS 2017. Although not directly comparable, this suggests that we can use our computed supervised scores as good state-of-the-art reference to compare our results to.

We did only scratch the surface on the possible applications of our proposed formulation. Future work will include unaffected samples that are actually healthy. Furthermore, the model architecture could be drastically simplified using one discriminator for both directions, allowing for larger generator networks as well as using multiple discriminators at different scales to find inpaintings that are not just locally but also globally consistent with the image. A restriction to slices is unfortunate but necessary due to memory requirements. A generalisation of our approach to volumetric data would make it feasible for more real clinical applications.
\paragraph{Conclusion}
We presented a new generative pathology segmentation model capable of handling a plethora of tasks: First and foremost, we presented a weakly supervised segmentation method for pathologies in 2D medical images, where it is only known if the image is affected by the pathology and thus no pixel-wise label or classification is provided. Furthermore, we were able to sample from both our healthy as well as our pathology model. We showed qualitatively and quantitatively, that we are able to produce compelling results, motivating further research towards actual clinical applications of PathoGAN.
\section{Acknowledgements}
We are grateful to the MIAC corporation for generously funding this work.
\bibliographystyle{splncs03}
\bibliography{paper}
\end{document}


\maketitle
\section{Additional Visual Results}
Additional visual results were randomly selected with the command \lstinline{ls testfolder | shuf | tail -n 20} and are shown in Figs \ref{HGG-Brats17_TCIA_335_1-73} to \ref{HGG-Brats17_CBICA_AQY_1-69}. The first row in the figures shows the available sequences (FLAIR, T1CE, T1, T2) and the manual segmentation, the second the respective generated inpaintings and probability map and the last row shows on the left the four final translations for each sequence as well as the generated probability map using the supervised MDGRU\cite{mdgru} reference method. 

\minipagelayoutfigure{HGG-Brats17_TCIA_335_1-73}

\minipagelayoutfigure{HGG-Brats17_CBICA_AQY_1-70}

\minipagelayoutfigure{HGG-Brats17_CBICA_AAG_1-71}

\minipagelayoutfigure{HGG-Brats17_CBICA_ABO_1-67}

\minipagelayoutfigure{LGG-Brats17_TCIA_630_1-60}

\minipagelayoutfigure{HGG-Brats17_TCIA_278_1-83}

\minipagelayoutfigure{LGG-Brats17_TCIA_177_1-64}

\minipagelayoutfigure{HGG-Brats17_TCIA_471_1-76}

\minipagelayoutfigure{HGG-Brats17_TCIA_296_1-62}

\minipagelayoutfigure{HGG-Brats17_TCIA_460_1-91}

\minipagelayoutfigure{HGG-Brats17_TCIA_198_1-68}

\minipagelayoutfigure{HGG-Brats17_CBICA_AOO_1-77}

\minipagelayoutfigure{HGG-Brats17_TCIA_218_1-65}

\minipagelayoutfigure{HGG-Brats17_TCIA_208_1-78}

\minipagelayoutfigure{HGG-Brats17_TCIA_606_1-83}

\minipagelayoutfigure{HGG-Brats17_2013_4_1-94}

\minipagelayoutfigure{LGG-Brats17_TCIA_645_1-79}

\minipagelayoutfigure{HGG-Brats17_CBICA_ANI_1-63}

\minipagelayoutfigure{HGG-Brats17_TCIA_332_1-79}

\minipagelayoutfigure{HGG-Brats17_CBICA_AQY_1-69}
\bibliographystyle{splncs03}
\bibliography{paper}